\documentclass[runningheads]{llncs}

\usepackage[T1]{fontenc}
\usepackage{graphicx}
\usepackage{graphicx}
\usepackage{booktabs}

\usepackage[utf8]{inputenc}

\usepackage{color}

\begin{document}
\title{Using Large Language Models to Assess Teachers' Pedagogical Content Knowledge
%\thanks{Supported by organization x.}
}
%
%\titlerunning{Abbreviated paper title}
% If the paper title is too long for the running head, you can set
% an abbreviated paper title here
%
\author{Yaxuan Yang\leavevmode\inst{1,2} \and
Shiyu Wang\leavevmode\inst{2} \and
Xiaoming Zhai\leavevmode\textsuperscript{*}\inst{1,3}}
\authorrunning{Yang et al.}
% First names are abbreviated in the running head.
% If there are more than two authors, 'et al.' is used.
%
\institute{
\leavevmode\inst{1}  AI4STEM Education Center, University of Georgia, Athens, GA 30602, USA \\
\leavevmode\inst{2} Department of Educational Psychology, University of Georgia, Athens, GA 30602, USA \\
\leavevmode\inst{3} Department of Mathematics, Science, and Social Studies Education, University of Georgia, Athens, GA 30602, USA \\
\email{Corresponding to: xiaoming.zhai@uga.edu}
}
%, Tiergartenstr. 17, 69121 Heidelberg, Germany
%\email{lncs@springer.com}\\
%\url{http://www.springer.com/gp/computer-science/lncs} \and
%ABC Institute, Rupert-Karls-University Heidelberg, Heidelberg, Germany\\
%\email{\{abc,lncs\}@uni-heidelberg.de}}
%
\titlerunning{Automatic Scoring of PCK}%
\maketitle              % typeset the header of the contribution

\renewcommand{\thefootnote}{\*}
\footnotetext{
\textbf{Cite the study:} Yang, Y, Wang, S., \& Zhai, X. (2025). Using Large Language Models to Assess Teachers’
Pedagogical Content Knowledge. Proceedings of the International Conference on AI in Education (Workshop), pp. 1-14, Palermo, Italy. }
\renewcommand{\thefootnote}{\arabic{footnote}}
\begin{abstract}
Assessing teachers’ pedagogical content knowledge (PCK) through performance-based tasks is both time and effort-consuming. While large language models (LLMs) offer new opportunities for efficient automatic scoring, little is known about whether LLMs introduce construct-irrelevant variance (CIV) in ways similar to or different from traditional machine learning (ML) and human raters. This study examines three sources of CIV—scenario variability, rater severity, and rater sensitivity to scenarios—in the context of video-based constructed-response tasks targeting two PCK sub-constructs: analyzing student thinking and evaluating teacher responsiveness. Using generalized linear mixed models (GLMMs), we compared variance components and rater-level scoring patterns across three scoring sources: human raters, supervised ML, and LLM. Results indicate that scenario-level variance was minimal across tasks, while rater-related factors contributed substantially to CIV, especially in the more interpretive Task II. The ML model was the most severe and least sensitive rater, whereas the LLM was the most lenient. These findings suggest that the LLM contributes to scoring efficiency while also introducing CIV as human raters do, yet with varying levels of contribution compared to supervised ML. Implications for rater training, automated scoring design, and future research on model interpretability are discussed.

\keywords{Construct-irrelevant variance (CIV)  \and Automatic scoring \and Pedagogical content knowledge (PCK)\and Large Language Models (LLMs)\and Artificial Intelligence (AI).}
\end{abstract}
\section{Introduction}
%p1. PCK is important but, is challenging to assess
%p2. Prior research of using machine learning for automatic scoring PCK; the limitations (labor intensive)
%p3. how this study addresses the research gaps. Research questions: (1) How much CIV is drawn by the variability of scenarios, the rater severity, as well as the judging sensitivity of the scenarios in the LLM scores on teachers' PCK? What are the differences in the variance between the two sub-constructs of PCK? (2) To what degree does the LLM scoring approach differ from that of the human scorers and traditional supervised ML scores with regard to the CIV?

Pedagogical Content Knowledge (PCK), defined as specialized knowledge that enables teachers to effectively transform subject matter knowledge into forms that are understandable and accessible to learners, plays an important role in effective teaching \cite{kind2009pedagogical}. As there is no clear agreement on its definition and many studies do not specify its components, it can be difficult to measure PCK \cite{depaepe2013pedagogical}. In recent years, PCK evaluation has been based on performance-based assessment and machine learning (ML) for automatic scoring \cite{wahlen2020automated}. However, supervised ML struggles to accurately score complex aspects of teachers' PCK, and training effective models requires large amounts of human-scored responses, making the development process resource-intensive \cite{zhai2020evaluation}. Compared to traditional ML approaches, Large Language Models (LLMs) can quickly adapt to new tasks like automatic scoring without the need for large, labeled training datasets, making it cost-efficient and time-efficient \cite{wu2025unveiling}.

Despite their potential, LLMs have shown promise for efficient automatic scoring, but it remains unclear whether they can accurately assess complex, context-dependent constructs such as teachers' PCK, especially in ways that align with human scoring logic and uphold validity. One important threat to score validity is construct-irrelevant variance (CIV), which refers to the portion of score variance that stems from factors unrelated to the construct being assessed \cite{messick1984psychology,haladyna2004construct}. CIV can arise from various sources in performance-based assessments, including inconsistencies in rater severity, differential rater sensitivity to task contexts, and variability in the scenarios used for eliciting responses \cite{zhai2021framework}. Specifically, in the context of PCK assessments, these sources of CIV may distort the interpretation of teacher performance by introducing unwanted variability that is not attributable to their actual pedagogical thinking \cite{zhai2020evaluation}.

Much of the existing research on automated scoring has focused on prediction accuracy or human-machine agreement, without examining whether the automatic scores are influenced by the same sources of CIV known to affect human scoring. To address this gap, this study examines LLMs' performance on automatic scoring of contextualized constructed responses of teacher PCK. It is guided by two research questions: (1) How much CIV is drawn by the variability of scenarios, the rater severity, as well as the judging sensitivity of the scenarios in the LLM scores on teachers' PCK? What are the differences in the variance between the two sub-constructs of PCK? (2) To what degree does the LLM scoring approach differ from that of the human scorers and traditional supervised ML scores with regard to the CIV?

\section{Assessing Teachers' Pedagogical Content Knowledge}
% categorize assessment methods (e.g., paper-pencil, observing teaching performance, alternative performance)
Various assessment methods have been employed to evaluate teachers’ PCK in educational research. One commonly used approach is the traditional paper-pencil assessment, which asks teachers to respond to written prompts in formats such as multiple-choice or short-response questions \cite{zhai2020evaluation}. This type of assessment is typically administered without contextual supports, such as classroom video scenarios. For example, some researchers developed a paper-pencil PCK assessment focused on the topic of photosynthesis \cite{park2018development}. The instrument presented teachers with written classroom scenarios followed by multiple-choice and open-ended items. Teachers were asked to identify student misconceptions and suggest instructional strategies in response to each scenario. This approach reflects a typical format of written-response PCK assessments, which do not incorporate contextual supports such as video or in-class observation.

Observation of teaching practice has also been used to evaluate teachers' PCK \cite{lee2007assessing}. It involved systematically observing teachers in the classroom multiple times throughout the school year. Researchers took field notes and collected related instructional documents to contextualize and interpret the teachers' instructional decisions and classroom practices. For example, some researchers conducted an observation-based assessment of pre-service science teachers’ PCK by analyzing their technology-integrated lesson plans and video-recorded microteaching sessions \cite{canbazoglu2016assessing}. Using a structured observation protocol, the researchers evaluated each teacher's instructional strategies, attention to student understanding, and use of curriculum materials. This method represents a typical approach to assessing PCK through observation by focusing on how teachers enact their knowledge in practice-based settings.

Furthermore, structured interviews have been conducted to elicit teachers’ thinking about lesson planning and classroom decision-making \cite{lee2007assessing}. These interviews included prompts regarding students’ prior knowledge, learning differences, and common science misconceptions, with the goal of capturing teachers’ knowledge of student learning and instructional strategies. For example, some researchers used structured interviews to explore how primary school teachers applied PCK when planning and teaching science lessons \cite{sothayapetch2013primary}. The interviews included questions about students’ prior knowledge, typical misconceptions, and approaches to supporting student understanding, aiming to reveal teachers’ PCK in the context of teaching electric circuits.

\section{AI and Large Language Models for Automatic Scoring}
% supervised machine learning vs. LLM for automatic scoring; meta-analysis
Supervised ML has been widely applied to automate scoring in educational assessment, particularly for constructed-response tasks. A recent meta-analysis \cite{zhai2021meta} synthesized 110 machine-human agreement (MHA) estimates and found that ML-based scoring systems can achieve moderate to substantial agreement with human raters, with a weighted average Cohen’s $\kappa$ of 0.64. However, the study also revealed significant variation in performance depending on specific features of the scoring system. Among the six examined moderators, algorithm type and subject domain had the largest effects: models using support vector machines (SVM) or sequential minimal optimization (SMO) achieved substantially higher agreement than those using regression or naïve Bayes classifiers, while assessments in biology yielded higher agreement scores than those in physics or chemistry. These findings highlight the sensitivity of supervised ML systems to both technical and domain-specific design choices, which can limit their generalizability and scalability.

In contrast, LLMs such as GPT-3.5 and GPT-4 offer a more flexible approach to automatic scoring. LLMs can perform scoring tasks with little or no labeled training data, enabling zero-shot or few-shot generalization across diverse domains and assessment types. Researchers demonstrated that GPT-4 can be prompted to generate high-quality synthetic responses for underrepresented scoring categories, thereby augmenting small datasets and improving the performance of scoring models without relying on additional student data \cite{fang2023using}.

Recent studies have also shown that LLMs can perform automatic scoring with precision comparable to traditional machine learning models and even human raters \cite{latif2024fine}. Unlike earlier approaches that rely heavily on labeled data and feature engineering, LLMs can generate scores and explanations with minimal task-specific training, often using zero-shot or few-shot prompts \cite{yang2025fine}. When fine-tuned or guided by well-constructed prompts, LLMs are able to align their outputs with human scoring rubrics, providing not only accurate predictions but also interpretable, rubric-based justifications \cite{lee2024applying}.

Moreover, prompt engineering strategies such as chain-of-thought reasoning and rubric conditioning have been shown to significantly improve scoring performance and transparency \cite{yang2025fine}. These methods help LLMs emulate human grading logic by generating intermediate reasoning steps before assigning scores. As a result, LLMs are emerging as powerful and flexible tools for automatic scoring across educational contexts, capable of scaling to new tasks while maintaining both accuracy and explainability \cite{latif2024fine}.

\section{Methods}
\subsection{Participants}
% secondary analysis of existing data; participants...
This study involved a secondary analysis of existing data collected from in-service science teachers \cite{zhai2020evaluation}. In the pilot phase, 192 teachers participated by responding to 11 classroom video clips. Based on pilot data, three clips were selected for use in the main study. A total of 187 teachers qualified for the testing phase. Among them, 12.3\% taught grades 3–5, 49.3\% taught grades 6–8, and 38.5\% taught grades 9–10. Three trained raters, named BD, AG, and ZB, independently coded the teacher responses. A consensus-based procedure was used to resolve discrepancies and ensure coding reliability.

\subsection{Instruments}
% videos, questions, scoring rubrics; human scoring approach, consensus-based; supervised machine learning approach and findings.

In the main study, each participating teacher viewed all three video scenarios and responded to two open-ended prompts per scenario. The first prompt asked, “What do you notice about the student ideas related to the science content in this video?” and the second prompt asked, “What do you notice about how the teacher responds to student ideas related to the science content in this video?” These prompts were designed to capture both teachers’ ability to analyze student thinking (Task I) and their understanding of teacher responsiveness (Task II). Each teacher provided six written responses in total, three for each type of task.

To evaluate open-ended responses, task-specific scoring rubrics were developed. Three trained raters independently coded all responses using these rubrics. A consensus-based approach was used to resolve any discrepancies among the raters, ensuring scoring consistency and construct validity.

Prior studies have applied supervised machine learning to score similar video-based PCK tasks, using human-labeled responses as training data \cite{zhai2020evaluation}. The machine achieved substantial agreement with human ratings ($\kappa = .805$ for Task~I; $\kappa = .570$ for Task~II). Compared to human raters, the machine was more severe but less sensitive to scenario differences, suggesting more consistent scoring across contexts.

\subsection{Sources of Construct-Irrelevant Variance}
This study focuses on three sources of CIV: scenario variability, rater severity, and rater sensitivity to scenarios, which are commonly observed in contextualized constructed-response assessments \cite{zhai2020evaluation}. These sources may introduce unwanted variation into scores that are not attributable to the target construct, thereby reducing the validity of score-based inferences.

Scenario variability refers to differences in the contexts and topics of the classroom video used to elicit responses. Even when designed to be equivalent, scenarios may vary in clarity, content familiarity, or instructional structure, influencing how teachers respond. Rater severity captures systematic differences in how strictly or leniently raters apply scoring criteria, regardless of response quality. Rater sensitivity to scenarios refers to variation in a rater's scoring behavior across different scenarios, such as being more lenient in one context and more severe in another. These sources were further modeled as distinct random facets in the multi-facet Rasch analysis, allowing us to estimate the degree to which each contributed to total score variance across tasks and rater types.

\subsection{Data Analysis}
\subsubsection{Prompting Large Language Models for Automatic Scoring.}
% prompt strategies... example
Recent research has shown that prompt design plays a central role in enabling LLMs to perform complex scoring tasks, particularly for open-ended scientific explanations \cite{yang2025fine}. In addition, several studies have compared different prompt engineering strategies for LLM-based scoring to find out how to better design the prompt. For instance, a recent study systematically compared six types of prompt configurations for GPT-based scoring using combinations of zero-shot and few-shot learning, chain-of-thought reasoning, and rubric-aligned context \cite{lee2024applying}. They proposed a structured process (WRVRT: writing, reviewing, validating, revising, and testing) to iteratively refine prompt effectiveness. Their findings showed that few-shot prompts with both contextual item stems and scoring rubrics yielded the highest scoring accuracy.

Based on previous studies, we designed structured prompts to guide GPT-4 in scoring teacher responses across analytic tasks. The objective was to determine whether each response demonstrated evidence of the targeted construct, as defined in the corresponding analytic rubrics for Task I and Task II. For each response, the model was asked to assign a binary score (0 or 1) and briefly justify its decision.

The prompt structure was consistent across both analytic tasks. Each system message began with a rubric restatement, followed by internal reasoning steps that instructed the model to detect relevant cues (such as references to student ideas or teacher responses) and verify alignment with scientific concepts or pedagogical moves. To support model reasoning, each prompt included three few-shot examples with paired binary labels and brief justifications. These examples illustrated how the scoring criteria should be interpreted in different contexts. While the reasoning steps were not explicitly returned in the output, they guided the model internally in making its classification. In addition, the prompts were customized for each type of task to reflect their distinct constructs.

To ensure consistent and analyzable outputs, we prompted the model to return two fields: a score and a brief explanation. This helped reduce variability in response format and enabled efficient large-scale analysis. The outputs were processed in batches, and the resulting scores informed our analysis of rater behavior and construct-irrelevant variance across scoring sources.

\subsubsection{Multi-facet Rasch Analysis.}
To examine score variability across multiple sources in an open-ended assessment context, we applied a multi-facet Rasch analysis via a generalized linear mixed model (GLMM), with the many-facet Rasch model (MFRM) defining the measurement structure and GLMM providing the estimation framework \cite{neittaanmaki2024all}. This approach models variation from different sources, including person ability, item difficulty, rater severity, and other contextual factors. Researchers have used GLMM to estimate MFRM structures while modeling complex interactions and accommodating binary outcome data typical of educational assessments \cite{kim2019analysis,neittaanmaki2024all}.

In our study, we specified random effects for four key facets: teachers’ PCK ability, scenario differences, rater severity, and rater-by-scenario sensitivity. These were estimated using a GLMM with a logit link function, appropriate for modeling dichotomous scoring outcomes in performance assessments \cite{greenwood2014scoring}. This framework allowed us to account for multiple sources of variability and context-specific rating patterns within a unified model.

We first organized the dataset by restructuring it from wide format to long format, with each row representing a single score assigned by a rater to a response. We included variables indicating the task, teacher ID, scenario, rater identity, and binary score. After preprocessing, we fitted separate GLMMs for Task I and Task II, treating them as distinct constructs. Each model included random intercepts to estimate the four facets of interest. The logit link function was used to model the probability of a 0 or 1 score. All models were fitted using the \texttt{glmer()} function from the \texttt{lme4} package in R.

To further investigate rater effects, we obtained the random intercept estimates for individual raters and for each rater-scenario combination. These values were interpreted as indicators of severity and sensitivity, respectively. We summarized these estimates to compare rater behavior across human raters, machine scoring, and language model scoring. We also used standard deviations, variance, and visual displays to describe the magnitude and consistency of these effects.

To evaluate model goodness-of-fit, we conducted a likelihood ratio test (LRT) comparing the full model to a reduced model excluding the random effects of interest. Variance components were extracted from the fitted models to assess the contribution of each facet. In addition, best linear unbiased predictors (BLUPs) were obtained for rater severity and rater-by-scenario interactions. These estimates were further summarized and visualized to examine differences in scoring patterns across rater types. Results are reported in the following section.

\section{Results}
\subsubsection{Model Fit.}

Table~\ref{tab:glmm_fit} summarizes the model fit statistics for the GLMMs used to analyze Task I and Task II. To evaluate model adequacy, we used log-likelihood values and LRT comparing each full model to a reduced model excluding random effects. Both models demonstrated significantly better fit than their corresponding null models, with $\chi^2(3) = 261.4$ for Task I and $\chi^2(3) = 3681.3$ for Task II, both $p < .001$. These results indicate that including variability associated with raters, scenarios, and their interactions significantly improved model fit.

\begin{table}[htbp]
\centering
\caption{Model fit summary and likelihood ratio test results for Task I and Task II GLMMs}
\label{tab:glmm_fit}
\begin{tabular}{lrrrrc}
\toprule
\textbf{Task} & \textbf{logLik} & \textbf{$\chi^2$ (df)} & \textbf{p-value} & \textbf{Converged} & \textbf{Singular Fit} \\
\midrule
Task I  & -2548.2 & 261.4 (3)   & < .001  & Yes & No \\
Task II & -2158.0 & 3681.3 (3)  & < .001  & Yes & Yes (scenario var = 0) \\
\bottomrule
\end{tabular}
\end{table}

Both models successfully converged. However, the Task II model resulted in a singular fit, with the scenario-level variance estimated as zero. This suggests that scenario differences contributed minimally to score variability in Task II.

\subsubsection{Research Question 1.}
To address our first research question, we examined the variance components from the GLMMs for each task. Table~\ref{tab2} summarizes the estimated variance for each facet. In Task~I, scenario-level variance was small (0.021), and in Task~II, it was estimated as zero, suggesting that the three scenarios in both tasks were comparable in difficulty and contributed minimally to score variability.

\begin{table}[t]
\centering
\caption{Variance components by facet for Task I and Task II models}
\label{tab2}
\begin{tabular}{llrr}
\toprule
\textbf{Task} & \textbf{Facet}          & \textbf{Variance} & \textbf{SD} \\
\midrule
Task I  & Teacher ID         & 11.050 & 3.324 \\
Task I  & Rater × Scenario   & 0.008  & 0.092 \\
Task I  & Rater              & 0.249  & 0.499 \\
Task I  & Scenario           & 0.021  & 0.146 \\
Task II & Teacher ID         & 7.081  & 2.661 \\
Task II & Rater × Scenario   & 0.129  & 0.360 \\
Task II & Rater              & 6.343  & 2.519 \\
Task II & Scenario           & 0.000  & 0.000 \\
\bottomrule
\end{tabular}
\end{table}

In contrast, the variance attributed to rater severity increased from 0.249 in Task~I to 6.343 in Task~II, and rater-by-scenario interaction variance (i.e., context sensitivity) rose from 0.008 to 0.129. These changes suggest that scoring variation introduced by raters became much more pronounced in Task~II, where the construct being assessed—teacher responsiveness—is likely more interpretive and less tightly defined than in Task~I.

\begin{figure}[htbp]
  \centering
  \includegraphics[width=0.9\textwidth]{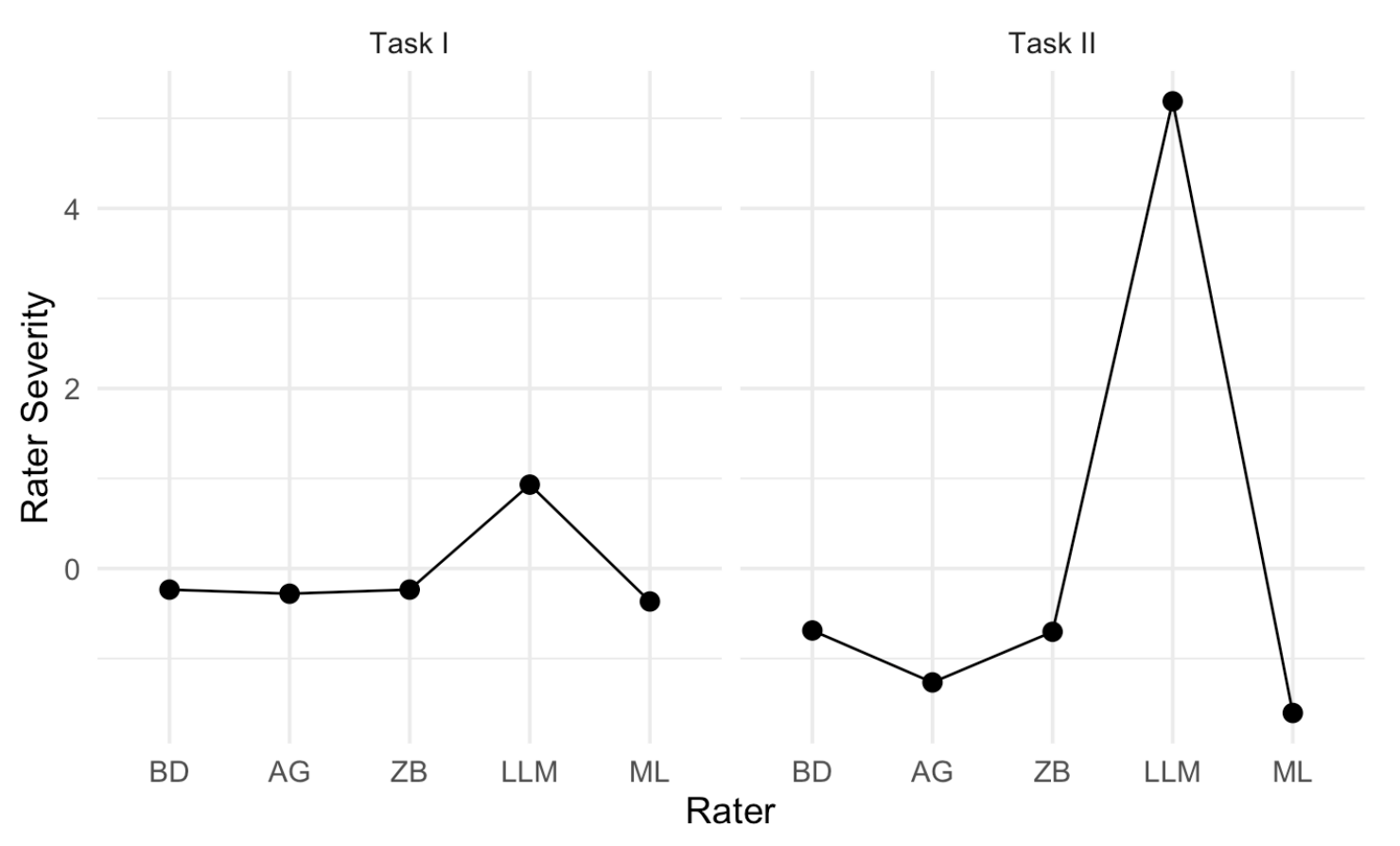}
  \caption{Dot plot of rater severity across Task I and Task II. LLM severity increased markedly in Task II, indicating a shift in rating behavior}
  \label{fig:severity-plot}
\end{figure}

To further unpack this rater-related CIV, we examined the rater-level random intercepts (Figure~\ref{fig:severity-plot}) estimated from the GLMMs. These intercepts represent each rater’s scoring severity on the logit scale, where larger values indicate more lenient scoring. As shown in the figure, the LLM consistently had the highest severity values across both tasks, indicating that it was the most lenient rater. The ML model was the strictest, with the lowest severity estimates, while human raters were clustered between the two extremes. Notably, the difference in severity among raters was much larger in Task~II, reflecting greater divergence in how raters interpreted and applied the scoring criteria in that task.

\begin{figure}[htbp]
  \centering
  \includegraphics[width=\textwidth]{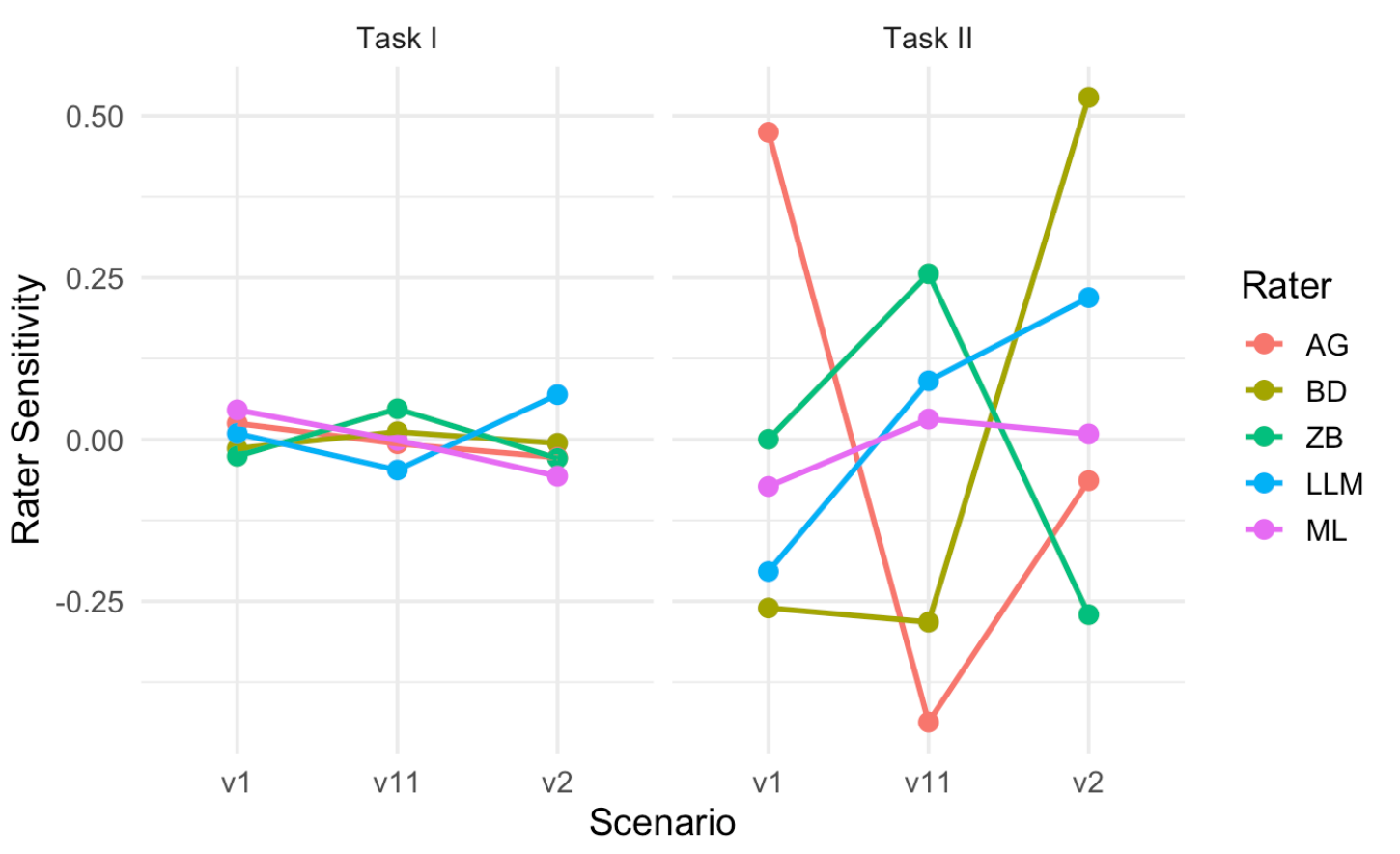}
  \caption{Rater-by-scenario sensitivity across tasks. Human raters showed greater variation across scenarios in Task II, while the LLM and ML scorers remained more stable, suggesting lower context-driven CIV in machine-based scoring}
  \label{fig:sensitivity}
\end{figure}

Rater sensitivity across scenarios is shown in Figure~\ref{fig:sensitivity}, which plots the rater-by-scenario interaction estimates for each rater. These values reflect how much a rater’s scoring deviates across different scenarios. In Task~I, rater sensitivity values were tightly clustered around zero, indicating stable scoring across contexts. In Task~II, however, the spread of values was much wider, particularly for human raters, suggesting higher sensitivity to scenario-specific features. The LLM and ML scorers exhibited relatively stable patterns, although the LLM showed slightly increased variation compared to Task~I.

These results suggest that while scenario features themselves contributed little to score variation, rater-related factors, especially severity and scenario sensitivity, introduced substantial CIV. This effect was especially pronounced in Task~II, likely due to the more interpretive nature of assessing teacher responsiveness. The distinct scoring patterns across rater types further highlight that the choice of scoring method can have a significant impact on the variability and comparability of scores in performance-based assessments.

\subsubsection{Research Question 2.}
To address our second research question, we compared the scoring behavior of the ML model, the LLM, and human raters in terms of rater severity and rater sensitivity, both of which represent potential sources of CIV. Table~\ref{tab3} displays the estimated rater severity BLUPs from the GLMMs, while Figure~\ref{fig:sensitivity} shows the rater-by-scenario interaction effects (sensitivity) across the two tasks.

\begin{table}[t]
\centering
\caption{Severity BLUPs for each rater across the two tasks}
\label{tab3}
\begin{tabular}{lrrr}
\toprule
\textbf{Rater} & \textbf{Severity (Task I)} & \textbf{Severity (Task II)} & \textbf{Diff (II - I)} \\
\midrule
BD   & -0.23 & -0.69 & -0.45 \\
AG   & -0.28 & -1.27 & -0.99 \\
ZB   & -0.23 & -0.70 & -0.47 \\
LLM  &  0.93 &  5.19 &  4.26 \\
ML   & -0.37 & -1.60 & -1.24 \\
\bottomrule
\end{tabular}
\end{table}

In our GLMMs, severity values are defined on the logit scale. A larger severity estimate corresponds to a higher log-odds of assigning a full-credit score, and therefore indicates a more lenient rater. Conversely, more negative severity values reflect stricter raters who are less likely to award full credit. As shown in Table~\ref{tab3}, the LLM had the highest severity estimates in both tasks, indicating that it was the most lenient scoring source. The machine learning model, by contrast, showed the lowest severity values, suggesting that it was the strictest. Human raters fell in between these two extremes and showed relatively stable severity patterns across the two tasks.

In terms of scoring consistency across contexts, Figure~\ref{fig:sensitivity} presents the rater-by-scenario interaction effects for each task. These values reflect how much each rater's scoring varied depending on the scenario. In Task~I, rater sensitivity values were generally close to zero, indicating consistent scoring across the three scenarios. However, in Task~II, raters showed much more variability. The human raters displayed more fluctuation in scenario-specific scoring patterns, while the machine learning model and the LLM appeared relatively more stable, although the LLM showed a slight increase in context sensitivity compared to Task~I.

Taken together, these results suggest that machine scoring systems differ from human raters in their contribution to CIV. The machine learning model appears to be overly strict but relatively stable across contexts, while the LLM is more lenient overall and slightly more sensitive to scenario variation in Task~II. Human raters, by comparison, occupy a middle position but exhibit higher context sensitivity, especially when the task demands more interpretation. These patterns highlight how different scoring sources introduce different types of construct-irrelevant variance into performance assessments, which has direct implications for fairness and validity.

\section{Discussion}
Our findings point to several important implications about where CIV arises in performance-based assessments of teacher knowledge. First, while the variance explained by the scenarios themselves was limited, we observed clear differences in how raters responded to the same classroom context. This suggests that even when scenarios are carefully designed to be equivalent, raters may still interpret them differently. This highlights that performance on PCK assessments is shaped not only by teacher knowledge but also by how the task is contextualized. Future research should explore what kinds of scenarios best capture the intended construct and minimize context-driven variation in scoring.

Second, we found that rater severity varied much more in Task II than in Task I. This task focused on evaluating teacher responsiveness, which involves more interpretive judgment than identifying student thinking. This may explain why scores were more sensitive to rater identity and scenario presentation. These patterns indicate that scoring open-ended professional reasoning tasks can introduce substantial CIV. Addressing this will require both stronger rater training and statistical controls that can adjust for rating tendencies across raters and tasks. This is especially important if performance assessments are to support valid inferences about teachers’ knowledge and decision-making.

Third, we found that machine scoring approaches do not behave in the same way as human raters. In particular, the LLM was the most lenient scorer, while the traditional ML model was the most severe. This pattern is consistent with prior work showing that LLM-based scoring tends to yield higher scores and greater alignment with human judgments \cite{cai2025rank}. These differences likely reflect  differences in how the two systems process written responses. LLMs may apply scoring criteria more flexibly due to their broader semantic representations, whereas traditional ML models rely more on fixed features and patterns. However, we still do not fully understand why these differences emerge or how they relate to task or item features. Future research should examine the mechanisms that drive severity and sensitivity in machine scoring, including how models interpret context, linguistic structure, and scoring prompts.

Finally, although automated scoring holds potential for expanding assessment capacity, it also raises interpretive and fairness concerns. Without transparency into how scoring decisions are made, it is difficult to know whether automated models are reinforcing construct validity or introducing new forms of CIV. Future work should not only continue improving the calibration of machine scoring but also investigate how model behavior aligns with the goals of performance assessment in professional education settings.

\section{Conclusions}
This study investigated how different scoring sources contribute to the CIV in the performance-based assessments of the PCK of teachers. Using a generalized linear mixed model framework, we examined three sources of CIV—scenario variability, rater severity, and rater sensitivity—across two analytic tasks and three scoring approaches: human raters, a traditional ML model, and an LLM.

Findings showed that while scenario variability was minimal, rater-related factors introduced substantial CIV, particularly in the task that required more interpretive judgment. The LLM consistently produced higher scores and appeared to be the most lenient rater, whereas the machine learning model was the most severe but also more stable across scenarios. Human raters fell in between, showing more variation in how they applied scoring across contexts.

These results suggest that scoring source matters for interpreting assessment outcomes and highlight the importance of monitoring and managing CIV in automated scoring systems. Future work should further explore how different models interpret open-ended responses and whether these patterns align with the intended constructs being assessed.

\subsubsection{Acknowledgements} This study secondarily analyzed material generated from prior work supported by the National Science Foundation under Grant No. DUE 1323162. We are grateful to the team members: Kevin Haudek, Christopher Wilson, Molly A.M. Stuhlsatz, Mark Urban-Lurain, John Merrill, Marisol Mercado Santiago, Michael Fleming, Zoë Elizabeth Buck Bracey, and Brian Donovan.

%
% ---- Bibliography ----
%
% BibTeX users should specify bibliography style 'splncs04'.
% References will then be sorted and formatted in the correct style.
%
% \bibliographystyle{splncs04}

%

%\begin{thebibliography}{8}

\renewcommand{\refname}{References}
 
\newpage

\bibliographystyle{splncs04}  
\bibliography{reference}

% \bibitem{ref_article1}
% Author, F.: Article title. Journal \textbf{2}(5), 99--110 (2016)

% \bibitem{ref_lncs1}
% Author, F., Author, S.: Title of a proceedings paper. In: Editor,
% F., Editor, S. (eds.) CONFERENCE 2016, LNCS, vol. 9999, pp. 1--13.
% Springer, Heidelberg (2016). \doi{10.10007/1234567890}

% \bibitem{ref_book1}
% Author, F., Author, S., Author, T.: Book title. 2nd edn. Publisher,
% Location (1999)

% \bibitem{ref_proc1}
% Author, A.-B.: Contribution title. In: 9th International Proceedings
% on Proceedings, pp. 1--2. Publisher, Location (2010)

% \bibitem{ref_url1}
% LNCS Homepage, \url{http://www.springer.com/lncs}. Last accessed 4
% Oct 2017
% \end{thebibliography}
\end{document}